\documentclass[10pt,twocolumn,letterpaper]{article}

\usepackage{3dv}
\usepackage{times}
\usepackage{epsfig}
\usepackage{graphicx}
\usepackage{amsmath}
\usepackage{amssymb}

\usepackage[pagebackref=true,breaklinks=true,colorlinks,bookmarks=false]{hyperref}

\threedvfinalcopy 

\def\threedvPaperID{****} 

\ifthreedvfinal\fi
\begin{document}

\title{SingleSketch2Mesh : Generating 3D Mesh model from Sketch}

\author{Nitish Bhardwaj\\
Accenture Labs\\
Bangalore, India\\
{\tt\small nitish.a.bhardwaj@accenture.com}
\and
Dhornala Bharadwaj\\
Accenture Labs\\
Bangalore, India\\
{\tt\small bharadwaj.dhornala@accenture.com}
\and
Alpana Dubey\\
Accenture Labs\\
Bangalore, India\\
{\tt\small alpana.a.dubey@accenture.com}
}

\maketitle

\begin{abstract}
Sketching is an important activity in any design process. Designers and stakeholders share their ideas through hand-drawn sketches. These sketches are further used to create 3D models. Current methods to generate 3D models from sketches are either manual or tightly coupled with 3D modeling platforms. Therefore, it requires users to have an experience of sketching on such platform. Moreover, most of the existing approaches are based on geometric manipulation and thus cannot be generalized. We propose a novel AI based ensemble approach, SingleSketch2Mesh, for generating 3D models from hand-drawn sketches. Our approach is based on Generative Networks and Encoder-Decoder Architecture to generate 3D mesh model from a hand-drawn sketch. We evaluate our solution with existing solutions. Our approach outperforms existing approaches on both - quantitative and qualitative evaluation criteria.
\end{abstract}

\section{Introduction}
As the world is shifting towards product customization, demand for 3D models is increasing at a very fast rate. The global 3D mapping and modeling market size is expected to grow to 6.5 billion by 2023 (CAGR of 18\%) \cite{3dstats}. This has made automated 3D model generation an important research topic in the field of computer vision \cite{IMSVR, fan2016point, groueix2018atlasnet, dvr, pixel2mesh++, 3dgan}. Many applications like robotics, animation, retail, virtual reality require precise and good quality of 3D models. With the recent development in 3D data sources and deep learning based approaches, there have been many successful attempts to generate good quality 3D mesh models from 2D colored images. However, Sketch to 3D generation still remains a challenging problem \cite{sketch_based_modeling}. The process of generating 3D model from sketch using CAD tools \cite{sketchbook, blender_grease, catianaturalsketch} requires high expertise and prior knowledge on these tools. Some of the AI approaches are limited by prior knowledge like camera calibration and multi-view information \cite{SketchModeling, pixel2mesh++, sketch_based_modeling}. 

\par In this paper, we propose a deep neural network based approach, SingleSketch2Mesh, to generate 3D mesh model from sketches. The approach is based on a two step process; in first step we generate multiple 2.5D views from a single sketch and in the second step we generate 3D model from these 2.5D views. 2.5D is an extended representation of 2D image or sketch with depth and normal information. Our approach uses deep neural network based models for each step. For sketch to 2.5D, we use autoencoder model with multi-view decoder. This model adds the missing depth information which is used by the next model. To generate 3D mesh model from multiple 2.5D views, we use GAN based architecture. This approach results in better quality of generated 3D mesh model from a sketch. The proposed system has been evaluated qualitatively and quantitatively. The evaluation results show that the proposed method generates significantly better quality 3D mesh models than the existing approaches. Moreover, as the approach neither requires any CAD tools expertise nor is limited by additional information like multiple views or camera parameters, it has much wider applicability. 
\par The main contributions of our work are:
\begin{itemize}
    \item We propose a solution to generate a 3D mesh model from a single view hand-drawn sketch. We put forward a training pipeline to train a model for any category object.
    \item We explore multiple ensemble approaches to generate 3D mesh models from sketches and evaluate their performance. 
    \item We propose and implement an extension to IM-NET \cite{IMSVR} network architecture to generate 3D representations from a single sketch. 
\end{itemize} 
\par The remainder of this paper is structured as follows: Section 2 discusses the related work on 3D object generation. The technical details of the proposed method are described in section 3. The experiments performed are discussed in section 4. We present the evaluation in section 5. Finally, section 6 concludes the paper with future work. 


\begin{figure*}
  \centering
  \includegraphics[width=14cm,keepaspectratio]{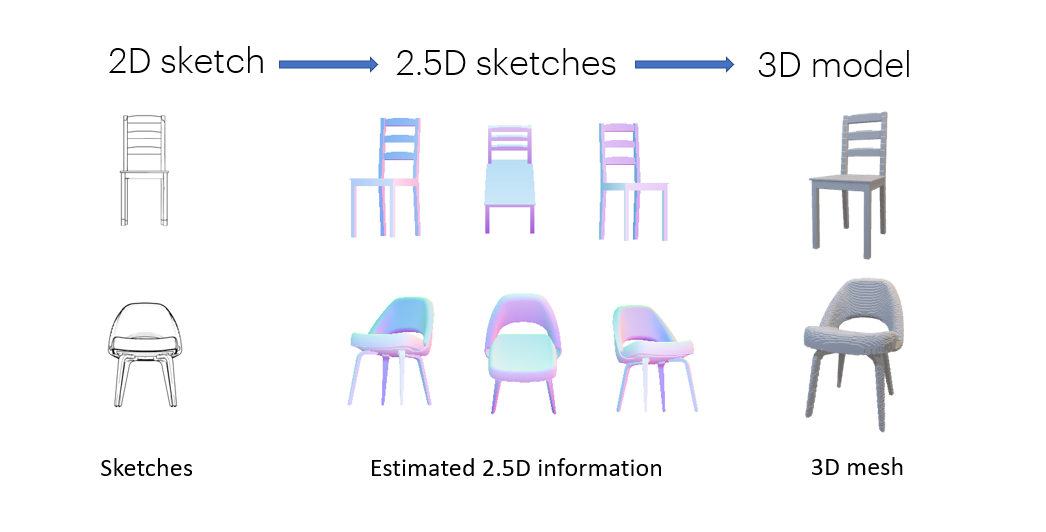}
  \caption{SingleSketch2Mesh : Converting sketch to 3D mesh model via 2.5D}
  \label{fig:SingleSketch2Mesh Framework}
\end{figure*}


\section{Related Work}
We present here the related work along three broad topics: first around CAD tools' capabilities to convert sketch to 3D model, second around deep neural network based approach for 3D object reconstruction, and third around sketch to 3D mesh model generation. \par
CAD tools like Autodesk, CATIA have developed software solutions to design 3D models from sketches. Autodesk's Sketchbook \cite{sketchbook} provides functionalities to designers to draw sketch with free hand and custom drawing tool. CATIA Natural Sketch \cite{catianaturalsketch} enables designers to sketch in CATIA 3D environment. This helps to avoid misinterpretations of multiple 2D views and supports in better visualization of their design ideas. Blender Grease Pencil tool \cite{blender_grease} gives flexibility to designers to sketch their idea easily and then export the sketch to blender 3D modelling tool to convert it to 3D model. The tools like FiberMesh \cite{fibermesh} and Teddy \cite{teddy} give sketching interfaces and a lot of flexibility to designers to use the tool and generate 3D model. FiberMesh \cite{fibermesh} presents a system for designing freeform surfaces with a collection of 3D curves. The designer can easily work with 3D curves as if working with a 2D line drawing. Teddy \cite{teddy} is based on geometric approach to extend 2D line sketch to 3D model. Sketching in these tools can easily help in 3D modelling but it requires high expertise and vast knowledge of the tool. These tools require professional artistic skills for sketching in 3D interface. Also, when CAD tool is used to convert sketch to 3D model, the generated model is not realistic as this process is based on geometric approach \cite{3D-from-sketch}. These tools can only work on new design concepts and can’t be used for existing sketches. These solutions are not scalable, i.e., it will be very time consuming to draw and create 3D models for a lot of designs.

\begin{figure*}
  \centering
  \includegraphics[width=16cm,keepaspectratio]{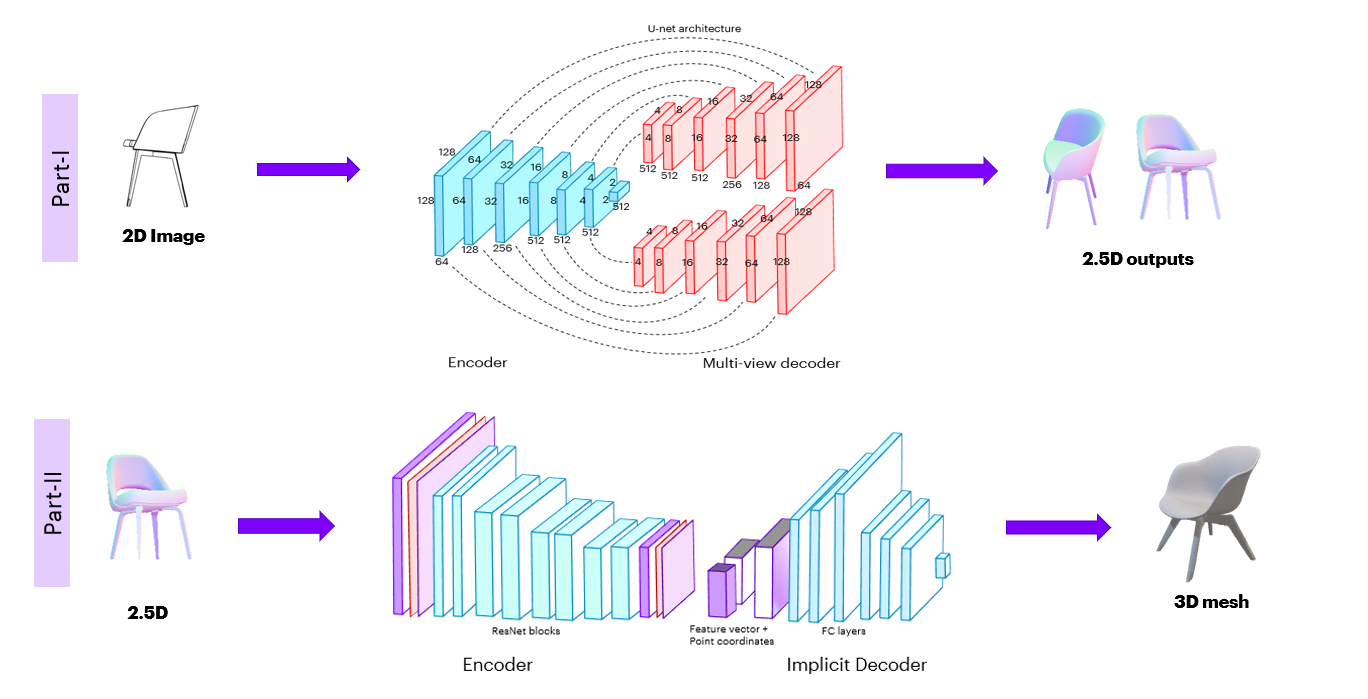}
  \caption{SingleSketch2Mesh Architecture}
  \label{fig:SingleSketch2Mesh Arhitecture}
\end{figure*}

\par
Recent development in Deep Neural network based solutions \cite{IMSVR, fan2016point, groueix2018atlasnet, dvr, pixel2mesh, pixel2mesh++, marrnet, 3dgan, shapehd, disn, genre} has shown better results in generating good quality 3D mesh models from 2D images. Some approaches combine geometric as well as neural network based approach to generate 3D mesh and some are purely based on deep neural networks. There are different 3D reconstructions methods for different 3D representations like voxel-based \cite{brock2016generative, choy20163dr2n2, gadelha20163d, rezende2018unsupervised, riegler2017octnet, wu2017learning, Xie_2019}, point-cloud based \cite{achlioptas2018learning, geometric_loss, Li_2018_ECCV, thomas2019kpconv, yang2019pointflow}, mesh-based \cite{groueix2018atlasnet, kanazawa2018learning, Liao2018CVPR, pan2019deep, pixel2mesh, pixel2mesh++} or implicit representations \cite{atzmon2019controlling, IMSVR, genova2019learning, Huang18ECCV, occupancy_networks, michalkiewicz2019deep, dvr, park2019deepsdf, disn}. Voxel can be easily represented mathematically and thus be easily processed by deep learning architectures, but this representation is limited by low resolution and memory inefficiency in up-scaling. Point clouds are memory-efficient but they lack connectivity information and thus require post-processing. Mesh-based approaches \cite{pixel2mesh, pixel2mesh++} have been very successful in generating good quality of 3D models but they still face the problem of missing patches thus leading to non-watertight meshes. Pixel2Mesh++ \cite{pixel2mesh++} takes images from different viewpoints along with their corresponding camera parameters and then generates a 3D model. In order to resolve these problems, there have been recent work on implicit representations \cite{atzmon2019controlling, IMSVR, genova2019learning, Huang18ECCV, occupancy_networks, michalkiewicz2019deep, dvr, park2019deepsdf, disn}. Implicit representation using occupancy networks \cite{occupancy_networks} allows to describe 3D geometry and texture implicitly. IMNET \cite{IMSVR} model takes images in multiple viewpoints and generates 3D mesh model using implicit-decoder \cite{IMSVR, occupancy_networks}. Existing solution like DISN \cite{disn} has similar approach to generate 3D mesh model using implicit network, but this approach works with RGB image, not Sketch. Other solution like DeepSDF \cite{park2019deepsdf} discusses a learned continuous Signed Distance Function (SDF) representation of shapes that enables high quality 3D reconstruction from partial and noisy 3D input data. This solution is based on encoder-decoder architecture, but it takes 3D data as input. \par
Although most of the work have been focused on converting image to 3D, there have been some work on sketch to 3D generation \cite{SketchModeling, marrnet, sketch_based_modeling}. As sketches are in gray-scale, they lack color details and thus it's difficult to extract depth information from hand-drawn sketches \cite{sketch_based_modeling}. SketchModelling \cite{SketchModeling} solution takes sketches in the aligned view and generates point cloud in various viewpoints which undergo fine-tuning and generates a mesh model. Marrnet \cite{marrnet} model generates 3D voxel models from 2.5D representation of sketches. However, the generated 3D models are not very refined. \par
Existing AI approaches take multiple views of sketches, camera parameters, viewpoint information and generate dis-oriented 3D models. In comparison to existing solutions, our approach uses only single sketch view to generate good quality of 3D model. Based on our observation about 2.5D representations, we develop an approach consisting of two deep neural network based models to generate 2.5D (sketch with depth information) and 3D model. So, if we compare our solution with existing approaches, our solution not only generates 3D models from sketches, but it generates 3D model from a single-view sketch. We believe that it is a significant improvement when compared with existing solutions.

\section{Workflow and Technical Details}

In this section, we first define some common terminologies we encounter in the model architecture. Next, we provide a detail description of our SingleSketch2Mesh approach and implementation details. 
\subsection{Common Terminologies}
In the paper, we use some of the terms, as follows:
\par
\textbf{2.5D representation:} 2.5D visual surface geometry is a representation that focuses on making a 2D space or image appear to have 3D qualities. These representations are surface projections of a 2D image onto a 3D environment making it an illusion in 2D profile. 2.5D information generally includes surface depth, normal and silhouette which are extended 2D projections of an image in 3D space. Earlier, animators achieved this look by strategically manipulating scenes with a set of tools such as layering, shadowing etc. in order to give the 2D static image a 3D animation \cite{cartoon-animator}. With the advent of neural networks and image processing, production of 2.5D information from a 2D image became easier. 2.5D depth provides us the distance of object from the viewpoint or camera co-ordinates while surface normal includes the normal projections of the image in 3D space.

\textbf{Occupancy networks:} Existing 3D data representations such as voxels, point clouds, etc. can not represent a 3D model efficiently due to the limitations such as memory restrictions in voxels and connectivity issues in point clouds. Though meshes excelled on voxel and point cloud representations with flexibility and connectivity but they cannot generate models handling multiple categories at same time, i.e. they are bound to the category they are trained on. With the ability to generate 3D models from any representation, occupancy networks \cite{occupancy_networks} became the new representation for learning-based 3D reconstruction. These networks represent a 3D geometry as a continuous function defined for every point $p \in \mathcal{R}^3$. 
\par The occupancy function  O : ${R}^3$ \verb|->| [0,1] moves in along with neural network assigning a probability value ranging between 0 to 1 for every possible point coordinate $p \in \mathcal{R}^3$. Occupancy networks output a decision boundary by setting an appropriate threshold value in the continuous function. These networks pose high representation power to reconstruct complex 3D shapes from either images, noisy point clouds or low resolution voxel representations when compared to state-of-the-art models. These networks upon applying smoothing algorithms produce a fine quality 3D objects.

\textbf{Implicit-decoder for shape generation:} Implicit-decoder \cite{IMSVR} varies from conventional decoders in producing high quality 3D objects. Unlike conventional approach which takes only feature vector extracted by a shape encoder as input, implicit decoder takes input a feature vector along with 2D or 3D point coordinate of the shape. The decoder outputs a probability value indicating the status of point relative to the shape. This decoder works on the concept of occupancy networks for predicting the status of the point lying inside or outside the 3D surface. More detailed description is followed in proposed approach.

\textbf{Marching cubes algorithm} \cite{marching_cube}: Existing 3D algorithms lack detail and sometimes introduce artifacts into objects. Marching cubes, a computer graphics algorithm is used for extracting a polygonal mesh of an iso-surface from a three-dimensional discrete scalar field (sometimes voxel). This algorithm works on generating the iso-surface around the discrete field endpoints by combining the points with connectivity concepts. Marching cubes algorithm has become one of the popular iso-surface extraction algorithms intended in preserving the triangle quality and topology correctness.

\subsection{Proposed approach}
Given a single view 2D sketch image, SingleSketch2Mesh generates 3D shape by following two parts (as shown in Figure \ref{fig:SingleSketch2Mesh Framework}): first, a 2.5D estimator predicts the depth and surface normal from the input sketch and generates multiple 2.5D views; second, a 3D implicit estimator infers 3D object shape in mesh representation using implicit function from the multiple 2.5D views.

\begin{figure}
  \centering
  \includegraphics[width=8cm,keepaspectratio]{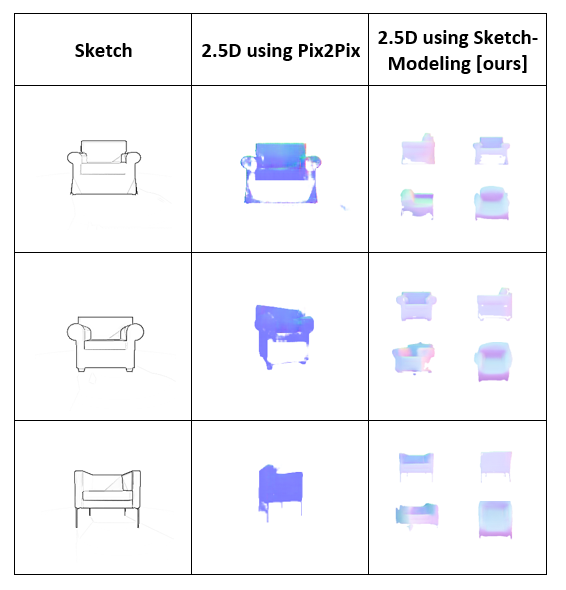}
  \caption{Sketch-to-2.5D}
  \label{fig:Sketch-to-2.5D}
\end{figure}

\subsubsection{\textbf{2.5D estimator from sketch:}} 
The working of the 2.5D estimator starts with a single-view 2D sketch image as input and multiple 2.5D outputs are generated for each input image. 2.5D outputs include depth and surface normal maps calculated for each input image when passed through the network. Earlier state-of-the-art techniques \cite{marrnet, shapehd, genre} perform 3D reconstruction using 2.5D as intermediate with encoder-decoder architecture. However, the recent work SketchModeling \cite{SketchModeling} proposed by Zhaoliang et al. performs 3D object generation using multiple refined hand-drawn sketches as input. It uses encoder-multiview decoder. SketchModeling first reconstructs a 3D point cloud from 2.5D representations; next it refines and fine-tunes multiple point clouds into a mesh model. A detail description about point clouds using Deep Learning can be found in \cite{point-cloud}.\par

In our approach, the encoder network consists of a series of convolutional layers followed by batch normalization and leaky ReLUs(with slope = 0.2) as activation function. All the filters are assigned a kernal size of 4 and stride 2. The output of this representation is 512 feature maps of size 2x2. The decoder network consists of a series of up-sampling and convolutional layers followed by batch normalization and leaky ReLU for each layer, but dropout for only first 3 layers. Decoder takes as input the encoder's representation and outputs a 256*256*5 image for a corresponding output viewpoint. The 5-channel image includes a depth map (1 channel), a normal map (3 channels) and a foreground probability map for that viewpoint. This model follows U-net architecture \cite{U-net}, the input of each convolutional layer in decoder is formed by combining the previous layer output in the decoder and corresponding layer output from the encoder. Network model predicts multiple 2.5D outputs from each vertex of a regular icosahedron with camera looking towards center of the object. Network parameters are updated to minimize the loss function by penalizing four terms (a) differences between the training depth maps and predicted depth maps, (b) angle differences between the training normal maps and predicted normal maps, (c) disagreement between groundtruth and predicted foreground masks, (d) large-scale structural differences between the predicted maps and the training maps. Let T represent training data consisting of sketches along with ground-truth foreground, depth and normal maps for V viewpoints, the loss function is expressed as
    \[L = \sum_{t=1}^T(\lambda_1 L_{depth}(t) + \lambda_2 L_{normal}(t) + \lambda_3 L_{mask}(t) + \lambda_4 L_{adv}(t)) \]
where $\lambda_1 = 1.0$, $\lambda_2 = 1.0$, $\lambda_3 = 1.0$ and $\lambda_4 = 0.01$ are weights tuned on validation set.
\par
\textit{ Depth and Normal loss}: Loss between predicted and ground truth is calculated using L1 distance for depths and cosine angle differences for normals. Let $S_t$ is a training sketch, $\hat{d}_{p,v,t}$ and $\hat{n}_{p,v,t}$ are ground truth depth and normal for the pixel p in viewpoint v. Each pixel is associated with a ground-truth label $ \hat{f}_{p,v,t} $, which is 1 for foreground, and 0 otherwise. The depth and normal predictions for the sketch $S_t$ are denoted as $d_{p,v}(S_t)$ and $n_{p,v}(S_t)$ respectively.
All the depths (i.e both training and predicted) are normalized within the range of [-1,1]. They are represented as 
\[L_{depth}(t) = \sum_{p,v}(|d_{p,v}(S_t) - \hat{d}_{p,v,t}|) \hat{f}_{p,v,t} \]
\[L_{normal}(t) = \sum_{p,v}(1 - n_{p,v}(S_t) \cdot \hat{n}_{p,v,t}) \hat{f}_{p,v,t} \]

\textit{ Mask loss}: Using cross-entropy function which is commonly used in classification, penalizing disagreement between predicted and ground-truth foreground labeling is performed.

\textit{ Adversarial loss}: Structural differences in the output maps are penalized with their corresponding ground truths through adversarial networks. The adversarial loss term takes input a 5-channel image I that contains the depth channel, the 3 normal channels, and foreground map channel and outputs the probability to be “real”
\[L_{adv} = -\sum_{v}\log(P(\frac{real}{I})) \]

The probability is predicted using the adversarial networks trained with ground-truth (real) maps $\hat{I}$ from generated (fake) maps I. The network in adversarial architecture is trained using the technique described in \cite{NIPS2014_5ca3e9b1}. 

\subsubsection{\textbf{3D implicit shape estimator:}}
To generate 3D mesh outputs, we require a network model to process 2.5D outputs generated from sketch estimator. The 3D shape estimator in our approach is defined as an implicit function F(p) which is continuous over 3D space. 


$$
F(p)=
\begin{cases}
1 & \text{if point is outside the shape,}\\
0 & \text{otherwise}
\end{cases}
$$

\par An implicit field is applied for each point $p \in \mathcal{R}^3$ predicting the inside/outside status of the point. It is similar to a binary classification problem which can be achieved using neural networks. A smooth surface extraction algorithm is implemented post implicit fields to yield a high quality 3D mesh model for various use cases.
\par
To finalize an architecture for 3D shape generation, there has been a comparison of  CNN decoders and implicit decoders in \cite{IMSVR}. Although CNN based decoder performed better in quantitative evaluation such as chamfer distance (CD), or global alignment, e.g., mean squared error(MSE) and intersection over union (IoU), but failed on qualitative metrics such as visual quality.

Inspired with the quality results of implicit decoders demonstrated for 3D shape generation \cite{IMSVR, dvr}, we use the similar architecture for 2.5D to 3D reconstruction. We enhance the architecture of implicit-decoder defined in IM-NET \cite{IMSVR} which has 6 fully connected layers. However, the decoder fails for single view reconstruction from 2.5D, as the network was over-parameterized. As mentioned in \cite{babaeizadeh2016noiseout}, we optimize the decoder with 5 fully connected layers and name this network as Extended IM-NET. 
\par
Similar to AtlasNET \cite{groueix2018atlasnet}, we first train an autoencoder, then update the parameters of the implicit decoder. In our experiments, we fix different parameters to minimize the mean square loss between the predicted feature vector and the ground truth. 

For our single-view 3D reconstruction from sketch, we follow ResNet \cite{RESNET} which contains residual blocks as an encoder. It is the best suited for feature extraction as the skip connections in ResNet solve the problem of vanishing gradient in deep neural networks by allowing an alternate shortcut path for the gradient to flow through. ResNet encodes images of shape 256*256 and outputs 128-dimensional feature vectors by minimizing the mean squared loss between the predicted feature vectors and the ground truth. Decoder architecture is designed as of the implicit decoder which takes feature vector extracted from encoder along with 2D/3D point coordinates corresponding to the shape as input and predicts the inside/outside field for each point as shown in figure \ref{fig:SingleSketch2Mesh Arhitecture}. These point coordinates includes point-value pairs extracted from voxelizing the 3D shape at different resolutions($16^3$, $32^3$, $64^3$, $128^3$).
\par
A sampling approach for generating point coordinates is used by taking the center of each voxel and generating $n^3$ points at different resolution. The loss function in 3D shape estimator is defined by the weighted mean squared error between ground truth labels and predicted labels for each point. Let S be a set of points sampled from the target shape having implicit field F. Let $w_p$ be the weight assigned to each point p with restricting implicit field in a unit 3D space, we attempt to find a function $f_\theta(p)$ with parameters $\theta$ that maps a point p $\epsilon [0, 1]^3$ to F(p). The loss function L is defined as:

   \[ L(\theta)  = \frac{\sum_{p \epsilon S} | f_\theta(p) - F(p)|^2 \cdot w_p}{\sum_{p \epsilon S} w_p}\]

We apply, marching cube algorithm \cite{marching_cube} on the generated 3D points values to generate a smooth high quality polygonal mesh.

\section{Experiments}
As our approach follows a two step process, we perform several experiments to compare the proposed approach with other state-of-the-art methods.
\subsection{Dataset:}
To train our network architecture, we need a dataset that includes training sketches, 2.5D images (depth and surface normal) along with corresponding 3D shapes. We use SketchModelling dataset \cite{SketchModeling}, which is a subset of ShapeNetCore dataset \cite{shapenet}, in our approach. It has 3 categories i.e. chair, airplane, and character each containing 3D object shapes along with hand drawn sketches and 2.5D information. As explained in Lun et al. \cite{SketchModeling}, the hand-drawn sketches in the form of line drawings are made by humans to convey shape information. The input sketches in the dataset is available in 3 views (top, side and front) with 4 variations for each input view, thus can be used for single-view and multi-view 3D object reconstruction. The training data also has 2.5D images for the sketches from 12 viewpoints, each taken from a regular icosahedron vertex. All training sketches and corresponding ground-truth depth and normal maps are rendered under orthographic projection according to the output viewpoint setting. Training of 2.5D sketch estimator is done using input sketches and corresponding 2.5D images, whereas training of 3D shape estimator is done using the 2.5D images for features and 3D shapes for point-value pair formations.

\subsection{Training Pipeline:}
There are two networks (part 1 and part 2), as shown in Figure \ref{fig:SingleSketch2Mesh Arhitecture}, and these are dependent on each other. Although, we perform training for both the networks separately, we use the same dataset preserving sketch to 3D mapping with the intermediate 2.5D. The output of training of part 1 generates 14 views of depth maps and normal maps for a single sketch. The 2.5D views represent vertices of regular icosahedron. The network learns the best view for 3D generation. For chair object, the slanted front view contributed the most and gives better output. We implement an implicit-decoder network to train model for 2.5D view to 3D generation.

\subsection{Testing Pipeline:}
SingleSketch2Mesh works with only one view of the test sketch. The trained model from part 1 takes single view of sketch as input and generates 14 views of depth maps and normal maps for the test sketch. We select the slanted front view and pass 2.5D representation to the trained model from part 2. This model generates 3D mesh model from 2.5D input. We have shown some test examples of Sketch and 3D pairs in Figure \ref{fig:figure_output_1}. 
\begin{figure}
  \centering
  \includegraphics[width=8cm,keepaspectratio]{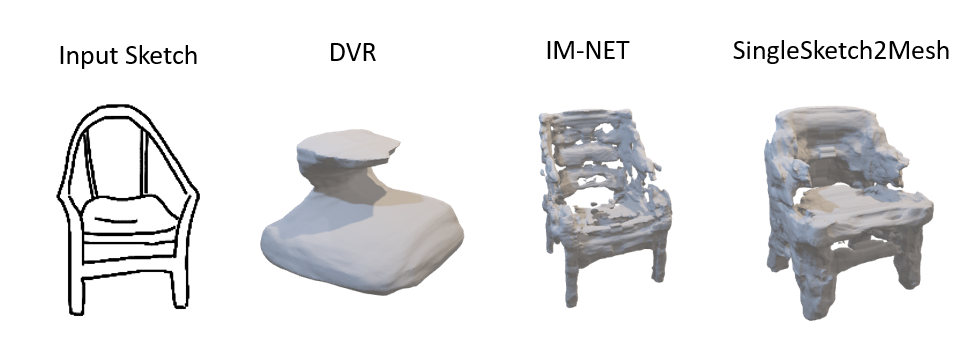}
  \caption{Comparison of state-of-the-art models for Single View Sketch to 3D Reconstruction} \label{fig:comparative_study}
\end{figure}
\subsection{Comparison with existing approaches:}
We perform comparative study to validate the importance of an ensemble model for 3D generation via 2.5D. The state-of-the-art models like DVR \cite{dvr} and IM-NET \cite{IMSVR} have shown great results for single view reconstruction for image and 2.5D. However, these models could not produce similar results for sketch as input. We use roughly drawn sketches, Sketchy \cite{sketchy2016} for validation of our solution. As shown in Figure \ref{fig:comparative_study}, the generated output of SingleSketch2Mesh is comparatively better than the other models. This validates our assumption that Image-to-3D reconstruction can't work for Sketch-to-3D generation. We conclude that the ensemble approach for Sketch-to-2.5D and 2.5D-to-3D can be a solution to solve this problem.
\par
As there does not exist an approach which generates 3D mesh model from a single view sketch, we are unable to compare it directly with existing approach. However, to evaluate the efficacy and usefulness of various 2.5D and 3D estimators for building an effective sketch to 3D mesh estimator pipeline, we conduct few experiments. We compare pipelines consisting of combination of Pix2Pix \cite{pix2pix}, DVR \cite{dvr}, IM-NET \cite{IMSVR} architectures as discussed in the next section. 
\par
\textbf{Pix2Pix} \cite{pix2pix} is an image-to-image translation network with conditional adversarial networks\cite{GAN}. The approach works on mapping one image to another, e.g., synthesizing photos from label maps, generating color maps from gray-scale and many other tasks. We use this pre-trained model to generate sketch-to-2.5D and sketch-to-image. 
\par
\textbf{Differentiable Volumetric Rendering} (DVR) \cite{dvr} is state-of-the-art model to generate 3D model from image. We use this pre-trained model for two intermediate steps, first 2.5D to 3D model and second Image to 3D model.
\par
\textbf{IM-NET} \cite{IMSVR} is another state-of-the-art model which uses the concept of occupancy network \cite{occupancy_networks} and implicit-decoder \cite{IMSVR} to convert image to 3D mesh model. In our approach, we extend the model so that it can work on single view sketch. 

\begin{figure}
  \centering
  \includegraphics[width=8cm,keepaspectratio]{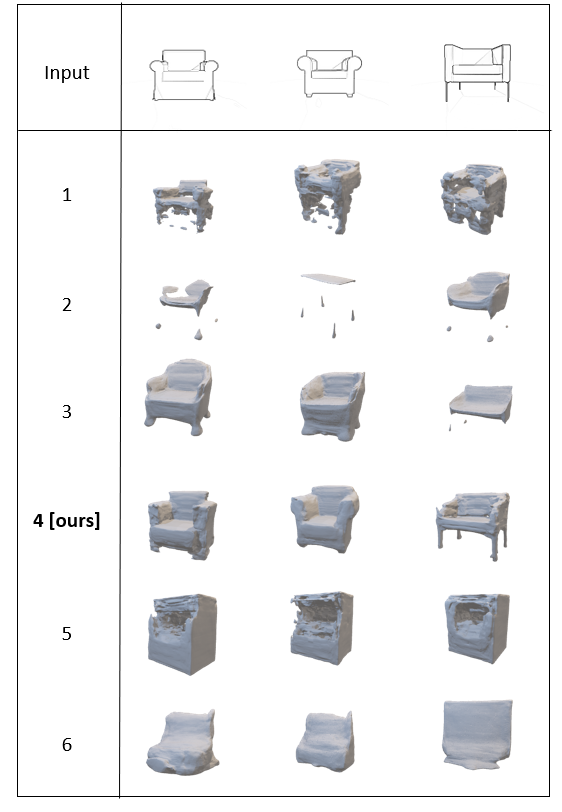}
  \caption{Qualitative Evaluation}
  \label{fig:Qualitative Evaluation}
\end{figure}
\par
\section{Results and Evaluation}
We evaluate our SingleSketch2Mesh model with state-of-the-art solutions. To the best of our knowledge, there are no known solutions for single view sketch to 3D generation. Therefore, we use an ensemble approach for creating a baseline. We compare different combinations of approaches for sketch to 2.5D  (as shown in Figure \ref{fig:Sketch-to-2.5D}) and 2.5D to 3D estimator (as shown in Figure \ref{fig:Qualitative Evaluation}). We evaluate approaches where 2D image and 2.5D are used as intermediate representations to compare the performance of such approaches. The proposed six approaches are as follows.


\begin{enumerate}
  \item 
    \begin{itemize}
     \item Sketch to 2.5D using Pix2Pix
     \item 2.5D to 3D using IM-NET
   \end{itemize}
  \item 
    \begin{itemize}
     \item Sketch to 2.5D using Pix2Pix
     \item 2.5D to 3D using DVR
   \end{itemize}
  \item 
    \begin{itemize}
     \item Sketch to 2.5D using Sketch-Modeling
     \item 2.5D to 3D using DVR
   \end{itemize}
  \item 
    \begin{itemize}
     \item Sketch to 2.5D using Sketch-Modeling
     \item 2.5D to 3D using Extended IM-NET
   \end{itemize}
  \item 
    \begin{itemize}
     \item Sketch to Image using Pix2Pix
     \item Image to 3D using IM-NET
   \end{itemize}
  \item 
    \begin{itemize}
     \item Sketch to Image using Pix2Pix
     \item Image to 3D using DVR
   \end{itemize}
\end{enumerate}
\par
Our proposed solution SingleSketch2Mesh follows the approach (4), Sketch to 2.5D using Sketch-Modeling and 2.5D to 3D using Extended IM-NET.
\begin{table}[t!]
  \centering
  \caption{Quantitative Analysis}
  \par
  \begin{tabular}{ |p{1cm}|p{1.5cm}|p{1.5cm}|p{1.5cm}|p{1.5cm}|}
    \hline \bfseries S.No. & \bfseries Sketch-to-2.5D & \bfseries 2.5D-to-3D & \bfseries Mesh Chamfer Loss & \bfseries Point Cloud Chamfer Loss\\ 
    \hline 1 & Pix2Pix & IM-NET & 7.660 & 0.018\\
    \hline 2 & Pix2Pix & DVR & 11.450 & 0.027\\
    \hline 3 & Sketch-Modeling & DVR & 11.542 & 0.037\\
    \hline 4 [ours] & Sketch-Modeling & Extended IM-NET & 4.982 & 0.018\\
    \hline \bfseries S.No. & \bfseries Sketch-to-Image & \bfseries Image-to-3D & \bfseries Mesh Chamfer Loss & \bfseries Point Cloud Chamfer Loss\\
    \hline 5 & Pix2Pix & IM-NET & 6.704 & 0.014\\
    \hline 6 & Pix2Pix & DVR & 10.682 & 0.024\\
    \hline
  \end{tabular}
  \label{table:1_evaluation}
\end{table}
\par
\par
\textbf{Quantitative Evaluation:}
We evaluate 6 approaches on two loss functions; mesh chamfer loss and point cloud chamfer loss (as shown in Table \ref{table:1_evaluation}). As we can see in Table \ref{table:1_evaluation}, SingleSketch2Mesh (approach 4) has shown better results than other approaches. Loss functions like mesh chamfer loss and point cloud chamfer loss for our approach are significantly less than the other models. 
\begin{figure*}
  \centering
  \includegraphics[width=14cm,keepaspectratio]{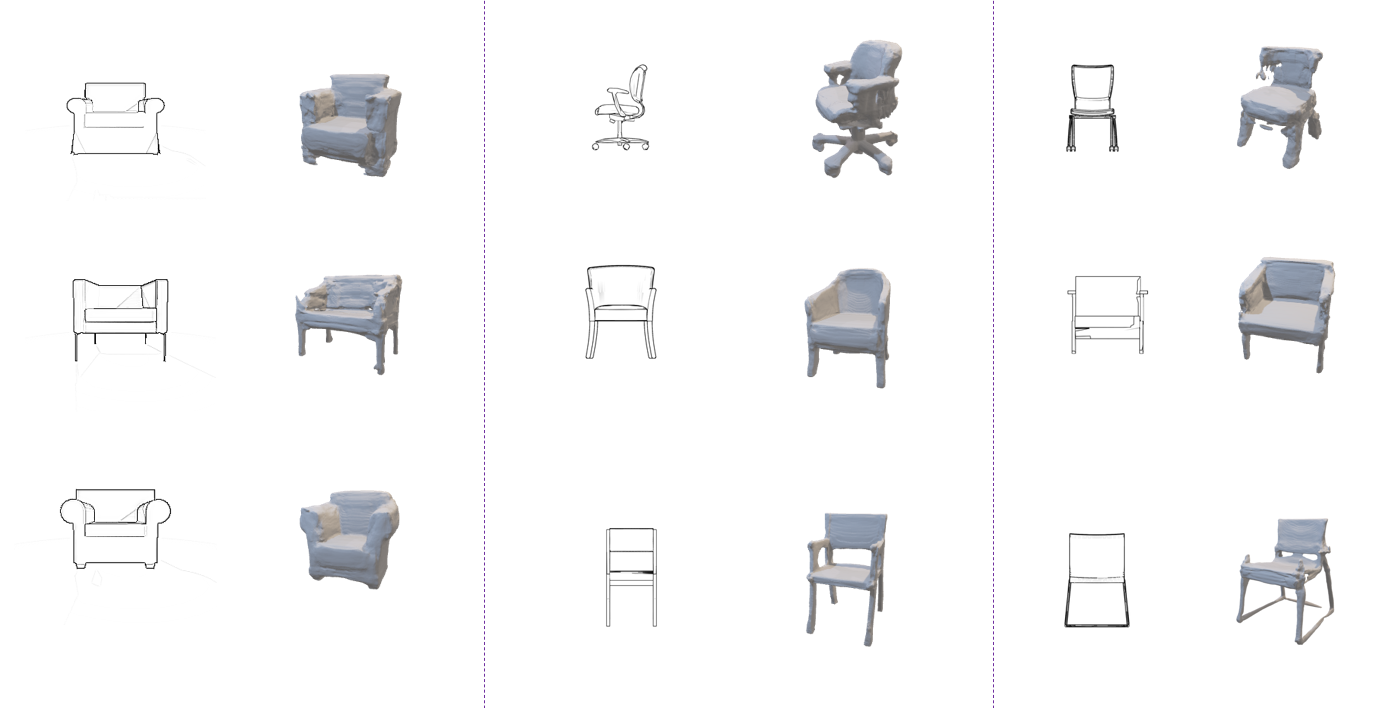}
  \caption{SingleSketch2Mesh Input-Output for User Study}\label{fig:figure_output_1}
\end{figure*}

\par
\textbf{Qualitative Evaluation:}
We compare the quality of generated 3D mesh models using six approaches (as shown in Figure \ref{fig:Qualitative Evaluation}). Our approach has resulted in more realistic and good quality 3D models. The outputs generated using 2.5D and DVR (approach 3) are very close to our solution but the resulted 3D model is not realistic as the surface has been flattened.

\begin{figure}
  \centering
  \includegraphics[width=6cm,keepaspectratio]{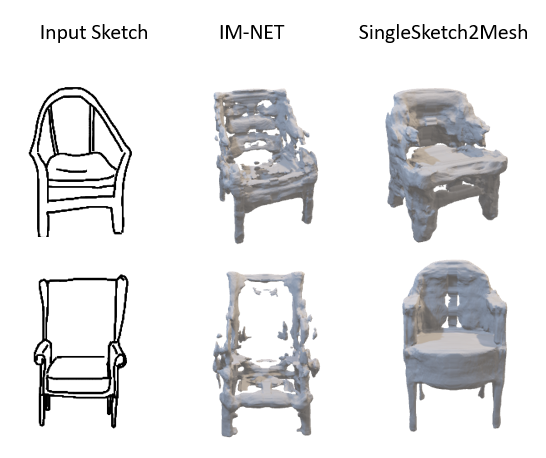}
  \caption{Ablation Study for Implicit-Decoder}\label{fig:ablation_study}
\end{figure}

\par
We conducted user surveys to evaluate the performance of our solution. We prepared 9 pairs of input sketch and generated 3D model (as shown in Figure \ref{fig:figure_output_1}). We presented users with following four questions. 1. Does generated 3D model represent the sketch? (Y/N) 2. Are there any feature of sketch missing in generated 3D model?(Y/N) 3. What is the quality of generated 3D model? (on a scale of 1-5)(5-best) 4. How close is the output with the input sketch? (on a scale of 1-5)(5-closest). 
\par
Some of the key findings from the user study are as follows. 78\% users agreed that the generated 3D model represent the sketch. As the 3D was generated from single-view of sketch, about 62\% noticed that some of the features were missing in the generated model. However, more than 80\% users rated the quality of generated output above average and about 85\% users also agreed that the generated 3D outputs were very close to the input sketches.
\par
\textbf{Ablation Study:}
We perform ablation study to evaluate optimization in implicit-decoder for 3D model generation. We compare the outputs of IM-NET with SingleSketch2Mesh which has used Extended IM-NET model as explained in section 3.2.2. As shown in Figure \ref{fig:ablation_study}, SingleSketch2Mesh generates 3D models with more details. 

\section{Conclusion and Future work}

In this work, we presented a solution to generate better quality 3D mesh model from a single view hand-drawn sketch via 2.5D generation. To the best of our knowledge, there does not exist any AI approach to generate 3D mesh model from single sketch. As the proposed solution SingleSketch2Mesh (approach 4) required input from single view, it worked more efficiently than mutli-view 3D shape generation (as shown in approach 3 \cite{SketchModeling, dvr}). Our approach also overcame the limitation of voxel-based approach \cite{marrnet} by generating high resolution of 3D mesh model. Moreover, we found that the ensemble approach to use 2.5D as an intermediate output was very useful in generating 3D models from single-view sketch. As it can work on single sketch, this solution has wider applicability as we may find situations where only singe view sketch is available. We believe that training the proposed model on a large dataset will make the solution more generic to work on any object. This solution can be integrated as plugin to any CAD tool. Designers are going to find this tool very helpful to visualize their initial design ideas into 3D models. \par
In the future, we plan to extend the approach to solid models so that it can be easily used in the CAD/CAM platform for further design and manufacturing activities. \par


{\small
\bibliographystyle{ieee_fullname}
\bibliography{main}
}





\end{document}